\documentclass{article}
\usepackage{iclr2026_conference,times}

\usepackage{hyperref}
\usepackage{url}
\usepackage{enumitem}
\usepackage{booktabs}
\usepackage{multirow}
\usepackage{subcaption}
\usepackage{graphicx}
\usepackage{algorithm}
\usepackage{algpseudocode}
\usepackage{amsmath,amssymb}
\algrenewcommand\algorithmiccomment[1]{\(\triangleright\)\,#1}
\usepackage{float}
\usepackage{wrapfig}
\usepackage{makecell}
\usepackage[table]{xcolor}
\usepackage{tabularx}
\usepackage{longtable}
\usepackage[most]{tcolorbox}
\usepackage{ltxtable}
\usepackage[numbers]{natbib}

\newcommand{\eg}{\text{e.g.}}
\newcommand{\ie}{\text{i.e.}}
\newcommand{\CFE}{\textsc{CFE-Bench}}
\newcolumntype{L}[1]{>{\raggedright\arraybackslash}p{#1}}

\definecolor{lightgray}{gray}{0.75}
\newcommand*{\email}[1]{\texttt{#1}}

\definecolor{primarybox}{RGB}{240,248,255}    
\definecolor{primaryline}{RGB}{70,130,180}    
\definecolor{secondarybox}{RGB}{255,250,240}  
\definecolor{secondaryline}{RGB}{210,105,30}  

\newenvironment{runningexample}[1][]{%
    \begin{tcolorbox}[
        colback=primarybox,
        colframe=primaryline,
        boxrule=0.5mm,
        sharp corners=all,
        before skip=10pt,
        after skip=10pt,
        left=4mm,
        right=4mm,
        top=2mm,
        bottom=2mm,
        breakable,
        title={#1}
    ]%
}{%
    \end{tcolorbox}%
}

\title{Classroom Final Exam: An Instructor-Tested Reasoning Benchmark}

\author{Chongyang Gao$^1$, Diji Yang$^2$, Shuyan Zhou$^3$, Xichen Yan$^4$, Luchuan Song$^5$, Shuo Li$^6$, Kezhen Chen$^6$ \\
$^1$Northwestern University, $^2$UC Santa Cruz, $^3$Duke University, $^4$University of Birmingham, \\ $^5$University of Rochester, $^6$Analogy AI, Inc. \\
\email{cygao@u.northwestern.edu}, \email{dyang39@ucsc.edu}, \email{shuyan.zhou@duke.edu}, \\\email{xxy315@student.bham.ac.uk}, \email{lsong11@ur.rochester.edu}, \\\email{shuoliqaq@outlook.com}, \email{kezhenchen@analogyai.org}\\
  \\
\vspace{-10mm}
}

\iclrfinalcopy 

\begin{document}
\maketitle

\begin{abstract}
We introduce \CFE{} (\textbf{C}lassroom \textbf{F}inal \textbf{E}xam), a multimodal benchmark for evaluating the reasoning capabilities of large language models across more than 20 STEM domains. \CFE{} is curated from repeatedly used, authentic university homework and exam problems, together with reference solutions provided by course instructors. \CFE{} presents a significant challenge even for frontier models: the newly released Gemini-3.1-pro-preview achieves an overall accuracy of 59.69\%, while the second-best model, Gemini-3-flash-preview, reaches 55.46\%, leaving considerable room for improvement. Beyond leaderboard results, we perform a diagnostic analysis by decomposing reference solutions into reasoning flows. We find that although frontier models can often answer intermediate sub-questions correctly, they struggle to reliably derive and maintain correct intermediate states throughout multi-step solutions. We further observe that model-generated solutions typically have more reasoning steps than those provided by the instructor, indicating suboptimal step efficiency and a higher risk of error accumulation. The data and code are available at \url{https://github.com/Analogy-AI/CFE_Bench}.\\
\\
\\

\textit{We are sincerely grateful to the following course instructors and professors—Eric Carlson, Yidong Chong, Jens Jensen, Kevin Zhou, Brian Naranjo, and Ravishankar Sundararaman—for explicitly approving and authorizing the use and release of their course materials for \CFE{}, and for verifying the corresponding problems and reference solutions.}
\end{abstract}

\section{Introduction}
Large language models and multimodal foundation models~\cite{gemma3,ministral3,llama4,gptoss,qwen3,qwen35,minimax21,minimax25,kimik2,kimik25,glm5,deepseek32,claude46,grok41,gpt52,gemini,internvl35} have advanced rapidly on a broad range of benchmarks. However, this progress has exposed a growing challenge: many widely used benchmarks are increasingly saturated, motivating the need for testbeds that are both more realistic and more discriminative~\cite{kiela2021dynabench,mcintosh2025inadequacies,ott2022mapping,owen2024predictable,rajpurkar2018know,glazer2024frontiermath}. At the same time, recent studies show that frontier models still struggle in advanced scientific and technical domains, particularly on problems that require deep domain knowledge and multi-step reasoning~\cite{phan2025humanity,wang2026frontierscience,yu2025hipho,feng2025physics}.

\begin{figure*}[t]
\centering
\includegraphics[width=1\textwidth]{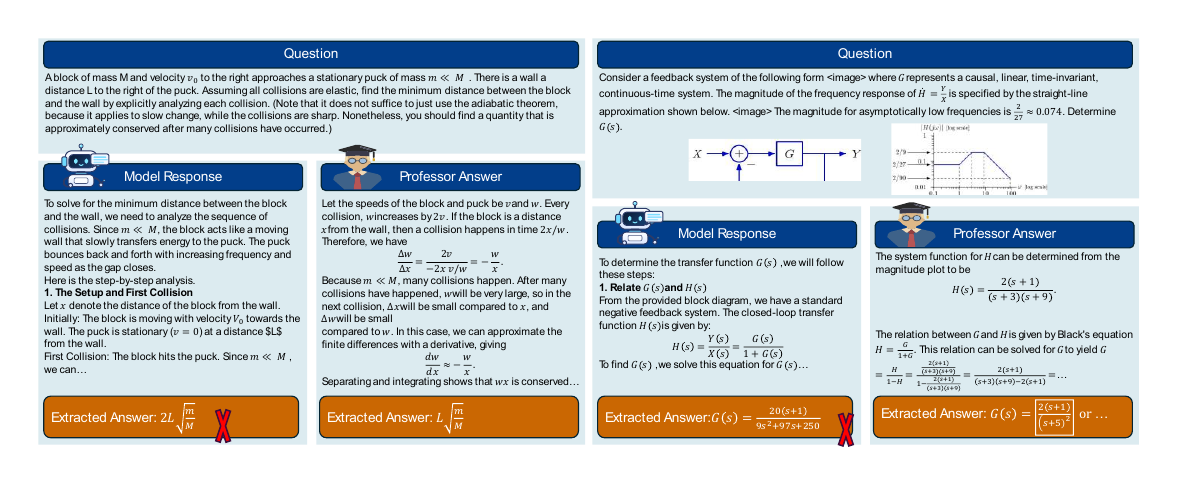}


\small
\centering
\begin{tabular}{p{0.5\textwidth}p{0.4\textwidth}}
\toprule
\textbf{Question} & \textbf{Variable-based Answer Annotation} \\
\midrule
A block of mass $M$ and velocity $v_{0}$ to the right approaches a stationary puck of mass $m\ll M$. There is a wall a distance $L$ to the right of the puck. Assuming all collisions are elastic, find the minimum distance between the block and the wall by explicitly analyzing each collision. 
& 
Variable: $x$ \newline Type: formula \newline Description: The minimum distance between the block and the wall, expressed as a mathematical formula. \newline Value: $L\sqrt{\frac{m}{M}}$ \\
\bottomrule
\end{tabular}

\caption{\textbf{Representative examples from \CFE{} and variable-based annotation.}
Top: example text-only and multimodal problems from \CFE{}.
Bottom: the structured annotation for the answer of the text-only example, including the variable name, type, semantic description, and ground-truth value.}
\label{fig:cfe_two_examples}
\end{figure*}

Motivated by these limitations, we introduce \textbf{\CFE{}} (\textbf{C}lassroom \textbf{F}inal \textbf{E}xam), a diverse text-and-multimodal reasoning benchmark sourced from authentic course materials maintained and verified by instructors.  Representative examples are illustrated in Figure~\ref{fig:cfe_two_examples}. 
\CFE{} comprises 449 high-quality problems partitioned into a text-only split (305 questions) and a multimodal split (144 questions). It covers more than $20$ Science, Technology, Engineering, and Mathematics (STEM) subjects, with substantial representation from Physics and Mathematics alongside multiple engineering disciplines and a long tail of additional domains, including computer science, chemistry, biology, and statistics.
\CFE{} is curated from repeatedly used advanced STEM questions from university instructors and established educational resources, providing strong reliability and classroom realism through prior instructional use and refinement.
To address challenges that real course materials often include open-ended questions or experiment-based requirements, we design \CFE{} with explicit selection and filtering criteria to ensure each item is (1) well-posed and objectively verifiable, (2) avoids trivial yes/no or multiple choice questions, and (3) does not require running physical experiments. 

Moreover, to reduce false positives from directly comparing long-form model responses with full reference solutions~\cite{thakur2025judging,jain2025consensusmitigatingagreeablenessbias,krumdick2025no}, as illustrated in Table~\ref{tab:gt_check_summary}, we introduce a more rigorous variable-based verification protocol. Specifically, we extract target answer variables from model outputs using annotated variable descriptions and types, and then compare the extracted variable values against the ground truth values, as illustrated in Figure~\ref{fig:cfe_two_examples}. Under this stricter evaluation, \CFE{} remains challenging: even the best-performing frontier model, the newly released \textbf{Gemini-3.1-pro-preview}~\cite{gemini}, achieves only \textbf{59.69\%} overall accuracy, while the best open-source model, \textbf{Qwen 3.5}~\cite{qwen35}, reaches \textbf{47.44\%}, leaving substantial room for improvement.
To further understand why frontier models fail on \CFE{}, we decompose reference solutions into reasoning units and conduct step-wise diagnostics of atomic competence, multi-step composition, and the impact of a single intermediate step. Across both text-only and multimodal settings, we find that strong models can often execute individual steps correctly when the sub-problem is specified, but they struggle to reliably derive and maintain correct intermediate states over long derivations. Notably, supplying a single correct intermediate answer can improve final-answer accuracy nearly as much as providing a long prefix of sub-questions, highlighting the importance of accurate intermediate states rather than decomposition alone. Finally, model-generated solutions typically exhibit longer reasoning flows than expert ground-truth solutions, indicating lower reasoning efficiency and creating more opportunities for intermediate errors to accumulate.

In summary, \CFE{} is built to provide the benchmark with a more reliable testbed for measuring underlying domain-grounded reasoning abilities under realistic academic standards. Our contributions are summarized as follows:
\begin{itemize}[leftmargin=*]
    \item \textbf{Benchmark:} We release \CFE{}, a reliable, unsaturated collection of diverse real-world classroom STEM problems, with both text-only and multimodal subsets.
    \item \textbf{Evaluation:} We propose a variable-based verification protocol for more accurate evaluation.
    \item \textbf{Diagnosis:} We introduce unit-based analyses that disentangle atomic execution from compositional failures and identify intermediate results that govern end-to-end success.
\end{itemize}

\section{Related Work}

\paragraph{Disparities in Reasoning Capabilities}
Recent evaluations demonstrate that Large Language Models achieve high performance on specialized reasoning tasks~\cite{huang2025winning}. Benchmarks such as MATH~\cite{lightman2023lets} and AIME~\cite{balunovic2025matharena} assess capabilities in competition-style mathematics. Furthermore, specific modalities are well-addressed by targeted benchmarks: OmniDocBench~\cite{ouyang2025omnidocbench} evaluates document parsing, while CharXiv~\cite{wang2024charxiv} focuses on complex chart interpretation. These benchmarks have established that current models can solve isolated, high-complexity problems or retrieve domain-specific facts with high accuracy.
However, strong performance on these targeted benchmarks does not necessarily imply systematic mastery of academic curricula. There remains a notable performance gap when models are presented with standard college-level coursework, which requires integrating vast domain knowledge (\eg, in-domain rules) with multi-step logical derivation.

\paragraph{Reasoning Benchmarks}
The validity of a reasoning benchmark is closely tied to its data source. Current approaches generally fall into two categories. The first relies on newly annotated or synthetic environments to ensure freshness and difficulty. Benchmarks like SimpleQA~\cite{wei2024measuring} and FACTS~\cite{kaggle-FACTS-leaderboard} push the boundaries of factuality but prioritize short-answer accuracy over the verification of long-form, compositional reasoning processes. Recent efforts, such as HLE~\cite{phan2025humanity}, address the need for complexity and multi-step tasks. However, as these solutions are newly authored or crowdsourced rather than time-tested, they remain susceptible to annotation errors and ambiguities that are often filtered out of established curriculum materials over years of use.

The second approach utilizes authentic, time-tested materials such as exams and textbooks. Benchmarks like ScienceQA~\cite{saikh2022scienceqa}, the MMLU series~\cite{hendrycks2020measuring,wang2024mmlu}, and the MMMU series~\cite{yue2024mmmu,yue2025mmmu} adopt this strategy to ensure pedagogical relevance. However, these benchmarks predominantly employ outcome-based metrics (\ie, multiple-choice accuracy), and recent evaluations with frontier models indicate that performance on these datasets is rapidly approaching saturation.
Furthermore, detailed explanations accompany only a small fraction of these questions. Even when available, the ``rationale'' provided is often post-hoc: designed to justify the answer's correctness rather than to demonstrate the constructive reasoning steps required to solve the problem. This limitation prevents such explanations from serving as intermediate checkpoints for model evaluation. 
CFE-Bench builds on authentic instructional materials and incorporates a stepwise evaluation framework. By leveraging expert-verified solution steps, the benchmark enables a diagnostic assessment of the model's reasoning flow.

\section{CFE Benchmark}
\begin{figure}[t]
    \centering
    \includegraphics[width=\linewidth]{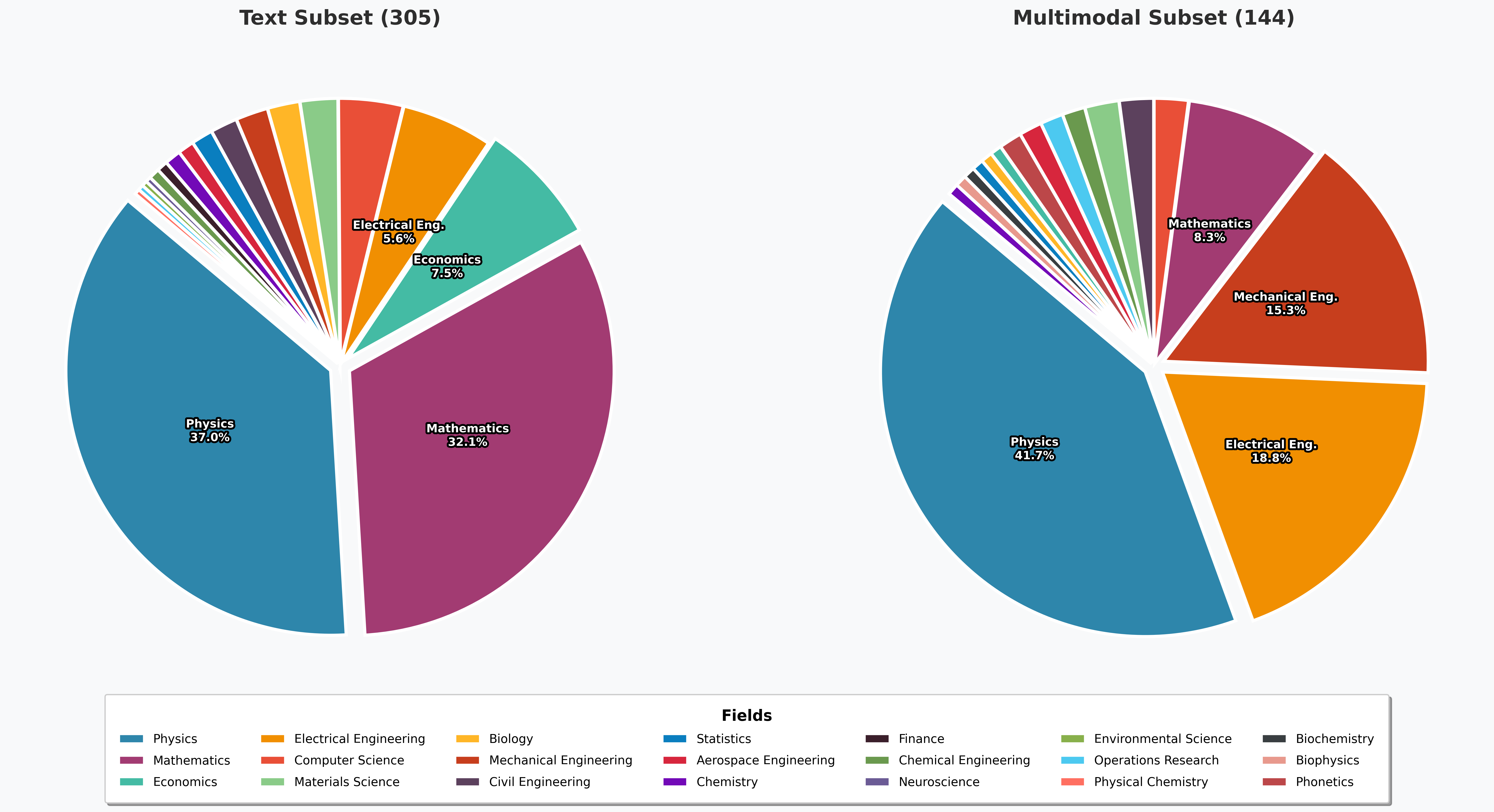}
    \caption{Subject distribution of CFE-Bench by modality.
    We report the field breakdown for the text-only subset ($305$; left) and the multimodal subset ($144$; right).}
    \label{fig:data_distribution}
\end{figure}

\subsection{Collection}
\label{sec:collection}
We curate problems from publicly available course resources, including exams, quizzes, and homework sets posted on instructor or course webpages.
Crucially, our collection emphasizes instructor-authored, classroom-tested materials: each item originates from a real course and is verified by domain experts and instructors for correctness and clarity.
Focusing on time-tested resources helps ensure that questions reflect realistic difficulty and have been validated through repeated instructional use and grading.
We prioritize problems that (i) require non-trivial multi-step reasoning, (ii) admit objectively checkable targets (\eg, numeric values, symbolic expressions, or well-defined outputs), and (iii) span both text-only and multimodal formats (\eg, diagrams, plots, circuit schematics, and geometric figures).
After collecting the raw questions, we perform extensive cleaning, including normalization of notations, standardization of units and symbols, and removal of duplicate or ambiguous items through a combination of LLM-assisted checks (\eg, similarity comparison) and expert review.
Each problem is then reviewed by human experts with relevant domain background.
Experts review each problem statement, filter out overly simple items, confirm that the target answer is well-defined, and ensure that the solution is verifiable under our evaluation protocol.
For the vision-language samples, we apply an additional filtering step to isolate image-dependent reasoning. We exclude samples that are solvable without visual input, as well as samples for which the model correctly answers all diagnostic sub-questions defined in Section~\ref{sec:exp} without visual input.

CFE-Bench contains two subsets: a \textbf{text-only} split with $305$ questions and a \textbf{multimodal} split with $144$ questions.
Figure~\ref{fig:data_distribution} summarizes the subject distribution, and Figure~\ref{fig:cfe_two_examples} provides two representative examples of CFE-Bench problems together with model responses and instructor solutions.
The text-only subset is dominated by Physics ($113$) and Mathematics ($98$), with additional coverage in Economics ($23$), Electrical Engineering ($17$), and Computer Science ($12$), among others.
The multimodal subset is similarly dominated by Physics ($60$) and spans multiple engineering domains, including Electrical Engineering ($27$) and Mechanical Engineering ($22$), as well as Mathematics ($12$), with a long tail of other STEM fields.
Overall, the benchmark emphasizes cross-disciplinary college-level reasoning, with substantial representation from physics and engineering and complementary coverage across mathematics and applied domains.

\subsection{Expert Annotation Protocol}
\label{sec:humanannotation}

To ensure that each problem statement is meaningful, unambiguous, and evaluable under our protocol, we conducted a human expert review and annotation process. 
We recruited 17 expert annotators with graduate-level (Master’s or above) education, all fluent in English, and spanning multiple domains relevant to our benchmark. The expert team was distributed globally to increase diversity of perspectives and reduce region-specific bias in interpretation. Across the project, the experts contributed a total of 945 working hours, covering screening, revision, and quality assurance.

We implemented a dedicated annotation interface to standardize expert review and collect structured, sample-level feedback. For each problem instance, the interface supports three core decisions: (i) filtering out overly simple or trivial questions, (ii) confirming that the target answer is well-defined, and (iii) verifying whether the reasoning flow (Section~\ref{sec:exp}) is correct.
For multimodal samples, the interface additionally assesses image dependency, \ie, whether visual information is necessary for the final answer and for the associated reasoning flow.

After expert review, refinement, and filtering, we retain 305 text-only questions and 144 multimodal questions, each paired with a human-verified reasoning flow. This interface-based workflow reduces annotation variance, promotes consistent decision criteria across experts, and improves auditability, since each edit and verification decision is linked to a specific item and reviewer.

\subsection{Variable-Based Evaluation}
\label{sec:variable_verification}

\begin{table}[t]
\centering
\small
\setlength{\tabcolsep}{3pt}
\renewcommand{\arraystretch}{1.1}
\begin{tabularx}{\linewidth}{l *{7}{>{\centering\arraybackslash}X}}
\toprule
Setting & Acc (\%) & AUC & P & R & F1 & FN & FP \\
\midrule
S2S & $98.03 \pm 0.77$ & $0.98 \pm 0.01$ & $0.98 \pm 0.01$ & $0.98 \pm 0.01$ & $0.98 \pm 0.01$ & $4.75 \pm 1.79$ & $1.25 \pm 0.83$ \\
L2S & $96.72 \pm 0.40$ & $0.96 \pm 0.01$ & $0.97 \pm 0.00$ & $0.97 \pm 0.00$ & $0.97 \pm 0.00$ & $1.50 \pm 0.50$ & $8.50 \pm 1.50$ \\
L2L & $89.67 \pm 0.75$ & $0.89 \pm 0.01$ & $0.90 \pm 0.01$ & $0.90 \pm 0.01$ & $0.90 \pm 0.01$ & $11.00 \pm 2.55$ & $20.50 \pm 1.12$ \\
\bottomrule
\end{tabularx}
\caption{Performance summary across evaluation settings.}
\label{tab:gt_check_summary}
\end{table}

Evaluating STEM solutions requires the careful examination of variances: answers may be correct but expressed in different algebraic forms, buried in verbose explanations, or interleaved with extraneous reasoning.
More importantly, conventional evaluation typically directly compares the model-generated response with the full reference answer. This \emph{long-to-long} (L2L) comparison can overestimate model capability by introducing non-trivial false positives.
Because an L2L judge must assess an entire narrative, verification can be confounded by (i) \textit{partial-correctness illusion} (many correct intermediate statements but an incorrect final value), (ii) \textit{context-induced judge error}, where the judge simultaneously observes a long model response and the reference solution and may be misled by the extended context, \eg, matching on superficial semantic overlap or implicitly treating the reference as the ``intended'' final answer rather than strictly verifying the model's produced targets, and (iii) \textit{fluency bias}, where highly coherent rationales increase the likelihood of acceptance despite subtle algebraic or computational mistakes.

To obtain a reliable, fine-grained metric robust to surface-form variation, we introduce a structured verification framework based on variable extraction, which we denote as \textbf{Short-to-Short structured verification (S2S)}.
For each question, we annotate a set of ground-truth variables
\[
V_{\mathrm{gt}}=\{(v_1,d_1,x_1,t_1),\ldots,(v_n,d_n,x_n,t_n)\},
\]
where each tuple consists of a variable name $v_i$, a semantic description $d_i$, the target value $x_i$, and a type $t_i \in \{\texttt{numeric},\texttt{formula},\texttt{other}\}$; an example is shown in the bottom panel of Figure~\ref{fig:cfe_two_examples}.
Annotations are produced and verified by human experts.

Given a model-generated response, we prompt a judge model to extract the predicted values $\hat{x}_i$ corresponding to each annotated variable specification $(v_i, d_i, t_i)$.
Then, the judge compares each extracted prediction $\hat{x}_i$ against the ground-truth value $x_i$, conditioned on the variable name, and returns per-variable correctness. We mark a response as correct only if all variables are verified as correct. 
The prompts used for extraction and verification are provided in Appendix~\ref{apx:prompts}.
We use a single judge model (GPT-mini) for both extraction and verification to ensure consistent evaluation across methods and models, as its performance is sufficient for this purpose (Table~\ref{tab:gt_check_summary}).

We compare our structured \textbf{Short-to-Short (S2S)} protocol against two alternative evaluation settings that differ in both the form of the model output and the form of the reference used for verification:
\begin{itemize}[leftmargin=*]
    \item \textbf{Long-to-Short (L2S).} We verify the long-form response from the model against the annotated variable values given their names, types, and descriptions.
    \item \textbf{Long-to-Long (L2L).} An end-to-end protocol in which the model's long-form solution is compared directly against the full reference solution using the same judge model.
\end{itemize}

To validate our evaluation protocol, we use Gemini-3-flash responses on the text-only split and ask domain experts to annotate each response as correct or incorrect. We then apply three automatic evaluation settings (L2L, L2S, and our S2S) using the model response, the full reference solution, and the variable-based annotations, and compare their judgments against the expert labels. This allows us to measure evaluation accuracy and error patterns (including false positives) for each setting.
As summarized in Table.~\ref{tab:gt_check_summary}, our \textbf{S2S} setting achieves the strongest overall agreement with expert annotations and substantially reduces false positives relative to \textbf{L2L}. By anchoring verification to concrete, typed target variables, S2S provides a more conservative and discriminative measure of model capability than holistic long-form matching, while remaining fully automatic at scale.

Based on our variable-based annotations, we report two complementary metrics: \textbf{Variable Accuracy}, which captures partial progress by averaging the fraction of correctly predicted variables per question, and \textbf{Question Accuracy}, which counts a question as correct only when \emph{all} annotated variables are correct. Together, these metrics provide a more fine-grained and informative evaluation than a single end-to-end correctness score.

\section{Model Performance on CFE-Bench}
\label{sec:sota_performance}

\begin{table*}[t]
\centering
\small
\begin{minipage}[t]{0.48\textwidth}
\centering
\scriptsize
\setlength{\tabcolsep}{1.8pt}
\begin{tabular}{llcc}
\toprule
\multicolumn{4}{c}{\cellcolor{blue!15}\textbf{Text subset ($305$)}} \\
\midrule
& \textbf{Model} & \textbf{Variable Accuracy} & \textbf{Question Accuracy} \\
\midrule
\cellcolor{green!10} & Gemma-3-27B-it & 13.79\% & 9.84\% \\
\cellcolor{green!10} & Ministral-3-14B-Reasoning & 17.92\% & 13.11\% \\
\cellcolor{green!10} & Llama-4-Maverick & 24.53\% & 19.67\% \\
\cellcolor{green!10} & GPT-oss-120b & 41.15\% & 34.43\% \\
\cellcolor{green!10} & Qwen3-235B-Instruct & 37.41\% & 32.46\% \\
\cellcolor{green!10} & Qwen3-235B-Thinking & 39.31\% & 32.79\% \\
\cellcolor{green!10} & Qwen3.5-397B & \textbf{\underline{54.12\%}} & \textbf{\underline{48.52\%}} \\
\cellcolor{green!10} & MiniMax-M2.1 & 33.44\% & 27.54\% \\
\cellcolor{green!10} & MiniMax-M2.5 & 34.46\% & 28.52\% \\
\cellcolor{green!10} & Kimi-K2-Instruct & 24.62\% & 19.02\% \\
\cellcolor{green!10} & Kimi-K2-Thinking & 46.28\% & 39.02\% \\
\cellcolor{green!10} & Kimi-K2.5 & \underline{51.32\%} & \underline{43.93\%} \\
\cellcolor{green!10} & GLM-4.7 & 44.79\% & 39.02\% \\
\cellcolor{green!10} & GLM-5 & 47.24\% & 41.64\% \\
\cellcolor{green!10} & deepseek V3.2 (chat) & 48.08\% & 41.64\% \\
\cellcolor{green!10} & deepseek V3.2 (reasoner) & 50.07\% & 43.28\% \\
\midrule
\cellcolor{orange!10} & Qwen3.5-plus & 54.43\% & 48.20\% \\
\cellcolor{orange!10} & claude-sonnet-4.5 & 36.74\% & 29.51\% \\
\cellcolor{orange!10} & claude-opus-4.5 & 49.03\% & 41.97\% \\
\cellcolor{orange!10} & claude-opus-4.6 & 58.95\% & 52.79\% \\
\cellcolor{orange!10}\multirow{-4}{*}[-2em]{} & Grok-4-0709 & 53.24\% & 47.54\% \\
\cellcolor{orange!10} & Grok-4.1-fast-reasoning & 49.58\% & 43.61\% \\
\cellcolor{orange!10} & GPT-5.2 & 57.99\% & 51.15\% \\
\cellcolor{orange!10} & Gemini-3-flash-preview & \underline{66.02\%} & \underline{58.69\%} \\
\cellcolor{orange!10} & Gemini-3-pro-preview & 65.29\% & 58.03\% \\
\cellcolor{orange!10} & Gemini-3.1-pro-preview & \textbf{\underline{70.66\%}} & \textbf{\underline{64.92\%}} \\
\bottomrule
\end{tabular}
\end{minipage}
\hfill
\begin{minipage}[t]{0.48\textwidth}
\centering
\scriptsize
\setlength{\tabcolsep}{1.8pt}
\begin{tabular}{llcc}
\toprule
\multicolumn{4}{c}{\cellcolor{purple!15}\textbf{Multimodal subset ($144$)}} \\
\midrule
& \textbf{Model} & \textbf{Variable Accuracy} & \textbf{Question Accuracy} \\
\midrule
\cellcolor{green!10} & Gemma-3-27B-it & 6.83\% & 2.78\%\\
\cellcolor{green!10} & Llama-4-Maverick & 16.09\% & 9.72\% \\
\cellcolor{green!10} & InternVL3-78B-Instruct & 6.52\% & 2.78\% \\
\cellcolor{green!10} & InternVL3.5-GPT-OSS-20B & 3.81\% & 2.08\% \\
\cellcolor{green!10} & InternVL3-5-38B & 10.23\% & 5.56\% \\
\cellcolor{green!10}\multirow{-5}{*}[-2.5em]{} & InternVL3.5-241B-A28B & 10.76\% & 4.86\% \\
\cellcolor{green!10} & Qwen3-VL-32B-Instruct & \underline{18.99\%} & \underline{10.42\%} \\
\cellcolor{green!10} & Qwen3.5-397B & \textbf{\underline{52.50\%}} & \textbf{\underline{45.14\%}} \\
\cellcolor{green!10} & GLM-4.6v & 15.15\% & 7.64\% \\
\midrule
\cellcolor{orange!10} & Qvq-max & 9.58\% & 5.56\% \\
\cellcolor{orange!10} & Qwen3.5-plus & 52.32\% & 44.44\% \\
\cellcolor{orange!10} & claude-sonnet-4.5 & 27.04\% & 18.75\% \\
\cellcolor{orange!10} & claude-opus-4.5 & 38.50\% & 30.56\% \\
\cellcolor{orange!10} & claude-opus-4.6 & 43.99\% & 36.81\% \\
\cellcolor{orange!10}\multirow{-5}{*}[-2.5em]{} & Grok-4-0709 & 36.23\% & 29.17\% \\
\cellcolor{orange!10} & Grok-4.1-fast-reasoning & 32.73\% & 26.39\% \\
\cellcolor{orange!10} & GPT-5.2 & 51.17\% & \underline{43.75\%} \\
\cellcolor{orange!10} & Gemini-3-flash & \underline{56.31\%}& \textbf{\underline{48.61\%}} \\
\cellcolor{orange!10} & Gemini-3-pro-preview & \textbf{\underline{57.22\%}} & \textbf{\underline{48.61\%}} \\
\cellcolor{orange!10} & Gemini-3.1-pro-preview & 56.26\% & \textbf{\underline{48.61\%}} \\
\bottomrule
\end{tabular}
\end{minipage}
\caption{\textbf{Model performance on CFE-Bench.}
We report two complementary accuracy metrics for both the text-only and multimodal subsets. 
\textbf{Variable Accuracy}: For each question containing multiple annotated variables, we compute the proportion of correctly extracted variables, then average this proportion across all questions.
\textbf{Question Accuracy}: The proportion of questions for which all variables are correct.
The leftmost column uses color coding: \textcolor{green!50}{\rule{8pt}{8pt}} green indicates open-weights models and \textcolor{orange!50}{\rule{8pt}{8pt}} orange indicates proprietary models.
\textbf{\underline{Bold underline}} indicates the best performance and \underline{underline} indicates the second-best performance within each group (open-weights and proprietary).}
\label{tab:cfe_performance} 
\end{table*}

\begin{wraptable}{r}{0.48\textwidth}
\centering
\scriptsize
\setlength{\tabcolsep}{1.8pt}
\begin{tabular}{llcc}
\toprule
\multicolumn{4}{c}{\cellcolor{gray!15}\textbf{Text + Multimodal ($449$)}} \\
\midrule
& \textbf{Model} & \textbf{Variable Accuracy} & \textbf{Question Accuracy} \\
\midrule
\cellcolor{green!10} & Llama-4-Maverick & 21.82\% & 16.48\% \\
\cellcolor{green!10} & Qwen3.5-397B & \textbf{\underline{53.60\%}} & \textbf{\underline{47.44\%}} \\
\midrule
\cellcolor{orange!10} & Qwen3.5-plus & 53.60\% & 47.00\% \\
\cellcolor{orange!10} & claude-opus-4.6 & 54.15\% & 47.66\% \\
\cellcolor{orange!10} & Grok-4-0709 & 47.78\% & 41.65\% \\
\cellcolor{orange!10} & Grok-4.1-fast-reasoning & 44.18\% & 38.08\% \\
\cellcolor{orange!10} & GPT-5.2 & 55.80\% & 48.78\% \\
\cellcolor{orange!10} & Gemini-3-flash-preview & \underline{62.90\%} & \underline{55.46\%} \\
\cellcolor{orange!10} & Gemini-3-pro-preview & 62.70\% & 55.01\% \\
\cellcolor{orange!10} & Gemini-3.1-pro-preview & \textbf{\underline{66.04\%}} & \textbf{\underline{59.69\%}} \\
\bottomrule
\end{tabular}
\caption{Combined performance on CFE-Bench (Text + Multimodal).}
\label{tab:cfe_combined_performance}
\end{wraptable}

We evaluate both open-source and proprietary models, including the Gemma-3~\cite{gemma3}, Ministral-3~\cite{ministral3}, Llama-4~\cite{llama4}, GPT-OSS~\cite{gptoss}, Qwen-3 / Qwen-3.5 series~\cite{qwen3,qwen35}, MiniMax series~\cite{minimax21,minimax25}, Kimi series~\cite{kimik2,kimik25}, GLM series~\cite{glm5}, InternVL series~\cite{internvl35}, DeepSeek-3.2~\cite{deepseek32}, Claude series~\cite{claude46}, Grok series~\cite{grok41}, GPT-5.2~\cite{gpt52}, and Gemini series~\cite{gemini}. We adopt a chain-of-thought~\cite{wei2022chain} prompting strategy for answer generation.
For decoding, we use the recommended temperature from the model documentation when available; otherwise, we set the temperature to $0.7$. For reasoning models, we limit the \emph{thinking} generation at 16{,}000 tokens, while leaving the final answer generation unlimited. We use default values for all other inference hyperparameters. When possible, we prioritize the official API provided by the model developer.

Table~\ref{tab:cfe_performance} reports performance of a broad set of state-of-the-art models on CFE-Bench, evaluated with our variable-based protocol. 
Across both subsets, question accuracy is substantially lower than Var.\ Acc., highlighting that models frequently solve some required components while failing at least one variable.
On the text-only split, the strongest results are achieved by \textbf{Gemini-3.1-Pro-Preview}, attaining the best overall performance (question accuracy  0.65). It also opens a clear margin over other leading proprietary models, including GPT-5.2 (0.51), Claude-Opus-4.6 (0.53), and Grok-4-0709 (0.48).
Among open-weight models, the best-performing system is \textbf{Qwen3.5}, followed by Kimi-K2.5 and DeepSeek-Reasoner.

The multimodal split is more challenging for all models and generally amplifies the gap between open-weight and proprietary systems. The \textbf{Gemini-3} family achieves the strongest performance, while \textbf{Qwen 3.5} ranks second overall across both open-weight and proprietary models, achieving a question accuracy of 0.45. This suggests that the leading open-weight and proprietary models are relatively close to the multimodal frontier. However, performance drops sharply for other open-weight vision--language models, remaining around or below 0.10 question accuracy, indicating a substantial capability gap outside the top tier.

We provide combined text-and-multimodal results in Table~\ref{tab:cfe_combined_performance}. \CFE{} remains clearly unsaturated: the strongest model (\textbf{Gemini-3.1-pro-preview}) reaches only \textbf{59.69\%} Question Accuracy.
We find that the gap between \textbf{Variable Accuracy} and \textbf{Question Accuracy} is consistent across models (typically \(\sim\)5--7 points), indicating that many responses make partial progress but fail to produce a fully correct solution. 
Second, the combined table highlights a clear frontier tier, led by the \textbf{Gemini} family. At the same time, the best open-weight/open-source model (\textbf{Qwen3.5-397B}) remains competitive, reaching \textbf{47.44\%} Question Accuracy, close to several proprietary systems.
Overall, Table~\ref{tab:cfe_combined_performance} supports the central claim of this benchmark: strong performance on popular public benchmarks does not imply robust classroom-level STEM reasoning. \CFE{} exposes substantial headroom in \emph{strict}, \emph{multi-step}, and \emph{multimodal} settings, making it a useful testbed for measuring meaningful progress beyond benchmark saturation.

\section{Deconstructing the Frontier Model Performance Gap}
\label{sec:exp}
To understand why even strong models remain far from reliable, we focus on Gemini 3 Flash.
Our diagnosis is organized around three questions and analyzes only the instances it fails to solve end-to-end:
\textbf{(Q1)} Does the model fail due to missing reasoning/knowledge at the atomic level?
\textbf{(Q2)} Is failure primarily driven by multi-step reasoning and error accumulation?
\textbf{(Q3)} Does providing a single critical reasoning unit, \eg, a key fact or transformation) substantially increase the probability of reaching the correct final answer?

\subsection{Formalizing the Reasoning Flow}
To diagnose the sources of failure on \CFE{} problems, we represent each instance as a structured \emph{reasoning flow}.
Given a question and its ground-truth final answer, we decompose the reference solution into an ordered sequence of verifiable reasoning units
$R \;=\; [u_1,u_2,\dots,u_n]$.
Each unit is defined as a question--answer pair
$u_i \;=\; \langle u_i^q,\;u_i^a\rangle$,
where $u_i^q$ is a unit-level sub-question that isolates a single step (\eg, retrieving a domain fact, applying a formula, or performing a local derivation), and $u_i^a$ is its corresponding verifiable target answer. The prompt to construct reasoning flow is illustrated in Appendix~\ref{apx:reason_flow_prompt} and an example reasoning flow is shown in Table~\ref{tab:reasoning_flow}.
This representation enables step-wise evaluation, allowing us to (i) probe whether models fail due to \emph{atomic} deficits (inability to execute a single unit) or due to \emph{compositional} deficits (inability to chain otherwise-solvable units), and (ii) define controlled interventions that condition model outputs on partial reasoning states (Sections~\ref{sec:exp1}--\ref{sec:step_length}).

We instantiate $R$ using a two-stage pipeline.
First, we prompt a judge model (GPT-mini) to propose a candidate decomposition into units $\{u_i\}_{i=1}^{n}$.
Second, human annotators review the proposed units to ensure that each $u_i^q$ is unambiguous given the prior context, with a corresponding $u_i^a$ that is objectively checkable and correct.

\begin{figure}[t]
    \centering
    \includegraphics[width=10cm, height=4cm]{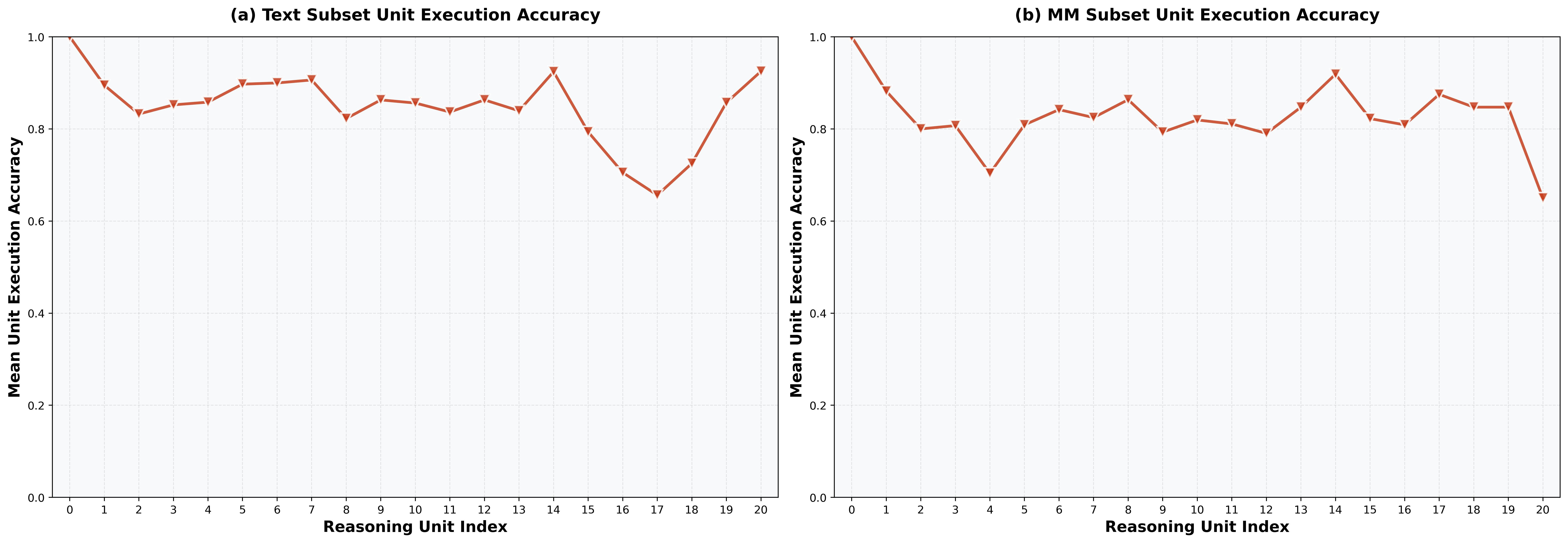}
    \caption{Unit Execution accuracy for text and multimodal subsets.}
    \label{fig:execution}
\end{figure}

\subsection{Q1: Unit Execution Ability}
\label{sec:exp1}
\paragraph{Setup.}
We isolate single-step competence via a unit execution test.
For each unit index $i$, we prompt the model with the original question $Q$, the preceding units $[u_1,\ldots,u_{i-1}]$, and the current sub-question $u_i^q$, and verify whether the model produces the correct unit answer $u_i^a$.
We run eight times per unit and aggregate outcomes by unit index, using the same generation settings as in Section~\ref{sec:sota_performance}.
We report, for each unit index $i$, the mean accuracy on sub-question $u_i^q$ across all questions whose reasoning flow includes unit $i$, averaged over repeated runs, as unit execution accuracy. To reduce variance from sparsely supported indices, we report step-wise averages only when more than five instances contribute to unit $i$.

\paragraph{Findings.}
Figures~\ref{fig:execution} show that \textbf{unit execution accuracy is consistently high} across most unit indices.
In the text subset, the model typically achieves $\sim$0.8--0.9 mean accuracy, and the multimodal subset exhibits a similar pattern with modest dips.
This indicates that many end-to-end failures are \textbf{not} explained by an inability to execute individual steps once the correct sub-question is specified.

\begin{figure*}[t]
    \centering
    \includegraphics[width=\textwidth]{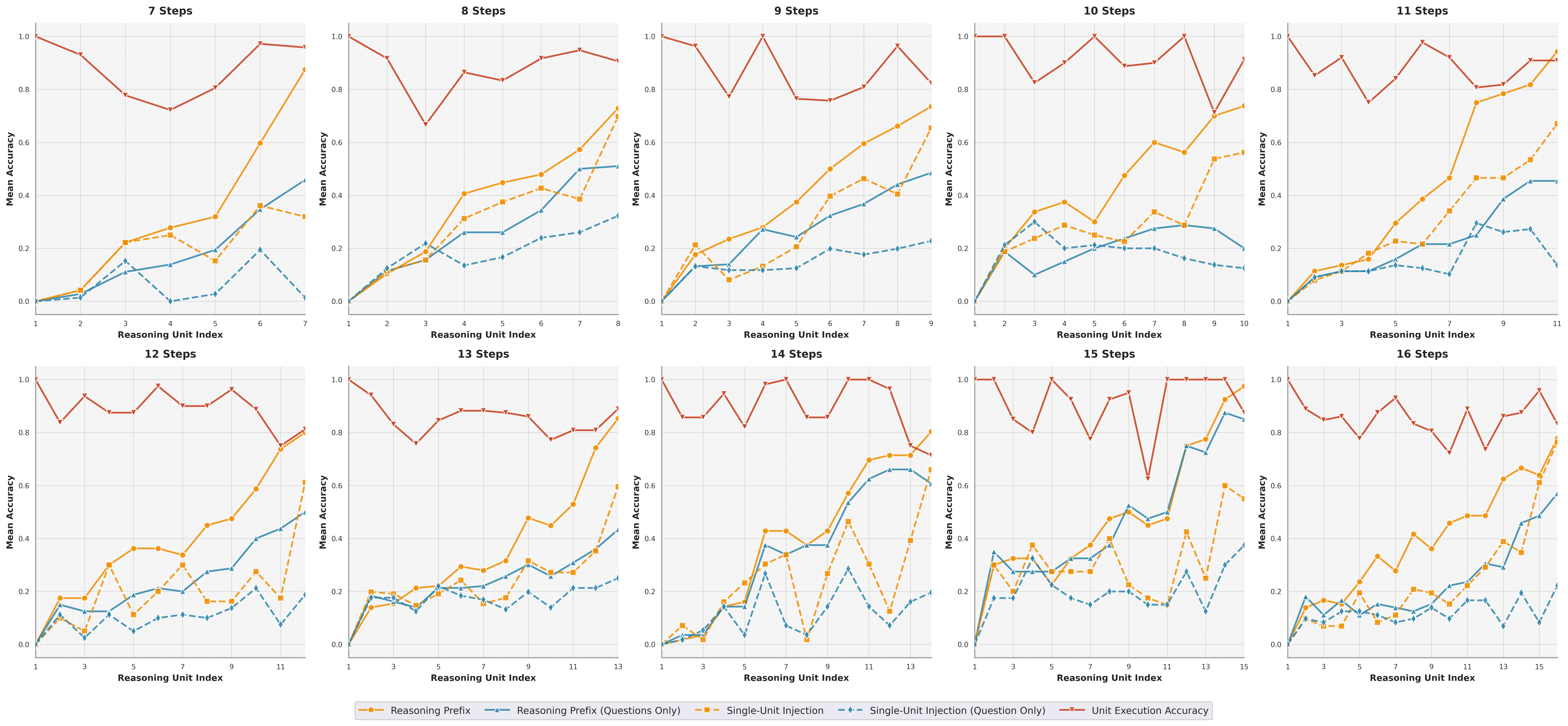}
    \caption{Sample-level diagnostics for text subset.
    The red curve shows unit execution accuracy.
    The other curves show final-answer accuracy under unit conditioning: \emph{Reasoning Prefix}, \emph{Reasoning Prefix (Questions Only)}, \emph{Single-Unit Injection}, and \emph{Single-Unit Injection (Question Only)}.
    Notably, although all curves share the same $y$-axis scale, the red curve measures \emph{unit-level} correctness, whereas the remaining curves measure \emph{final-answer} correctness.}
    \label{fig:text_step_conditioned_sample_level}
\end{figure*}

\begin{figure*}[t]
    \centering
    \includegraphics[width=10cm, height=6cm]{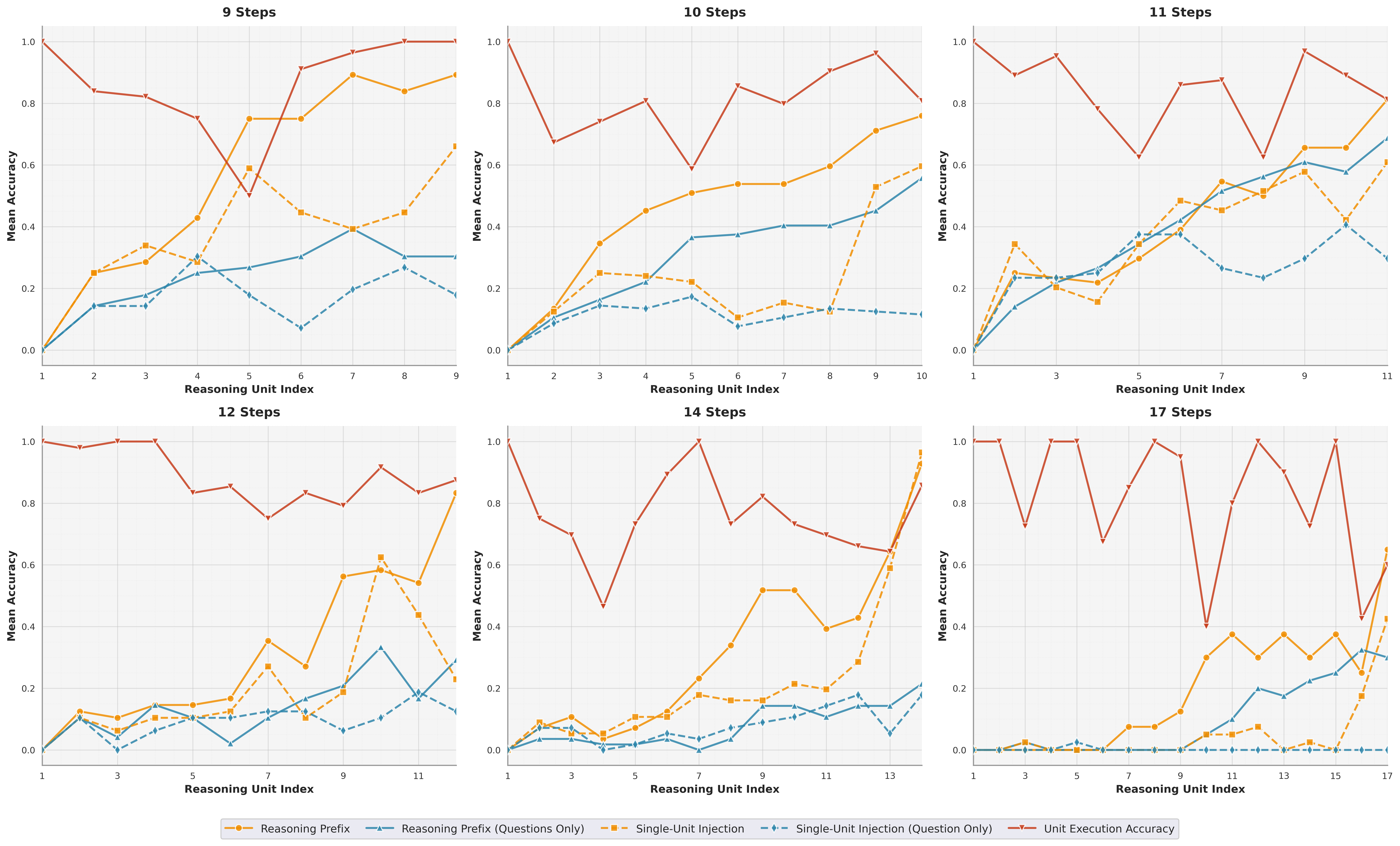}
    \caption{Sample-level diagnostics for multimodal subset.
}
    \label{fig:mm_step_conditioned_sample_level}
\end{figure*}

\subsection{Q2: Reasoning Progression Capability}
\label{sec:exp2}
\paragraph{Setup.}
To examine multi-step composition, we measure final-answer accuracy when progressively more of the reasoning flow is provided.
For each unit index $i$, we evaluate four prompting conditions: Reasoning Prefix\ ($Q+[u_1,\ldots,u_i]$), Reasoning Prefix (Questions Only) ($Q+[u_1^q,\ldots,u_i^q]$), Single-Unit Injection ($Q+u_i$), and Single-Unit Injection (Question Only) ($Q+u_i^q$).
For each condition and index $i$, we sample the model 8 times, compute per-question accuracy as the fraction of correct runs, and then average across questions.
For the sample-level analysis (Figures~\ref{fig:text_step_conditioned_sample_level} and~\ref{fig:mm_step_conditioned_sample_level}), we stratify questions by their total reasoning-flow length $s$ and report results only within buckets containing questions with exactly $s$ units.
We restrict attention to buckets with at least five questions, yielding $s\in\{7,\ldots,16\}$ for the text subset and $s\in\{9,10,11,12,14,17\}$ for the multimodal subset.

\paragraph{Findings.}
Across both modalities in Figures~\ref{fig:text_step_conditioned_sample_level} and~\ref{fig:mm_step_conditioned_sample_level}, providing unit answers consistently outperforms providing only unit sub-questions.
This gap indicates that the bottleneck is not merely identifying an appropriate decomposition, but \textbf{reliably constructing correct intermediate states}, \ie, producing the concrete intermediate values/expressions and preserving constraints as the derivation progresses.

Notably, the performance gap is largest at \emph{mid-range} unit indices.
Early units tend to involve local setup and relatively direct manipulations, where the model can often proceed even without answer supervision; later units are increasingly dominated by the hardest long-flow instances and require a precise final consolidation step.
In contrast, mid-chain units typically involve the most error-prone transformations, combining multiple prior results, applying the correct identity/theorem, and executing multi-step algebraic or numerical operations.
Providing the unit answers effectively ``bridges'' these difficult transitions, yielding the largest accuracy gains in the middle of the reasoning flow.
Overall, this pattern suggests that current SOTA models are comparatively better at using correct intermediate results once provided, but remain brittle at deriving them and at faithfully maintaining state over long STEM derivations.

\subsection{Q3: Critical Reasoning Units}
\label{sec:exp3}
\paragraph{Setup.}
We test whether end-to-end success depends on a critical intermediate unit by injecting only a single unit $u_i$ or only its sub-question $u_i^q$. We evaluate whether the model correctly answers the final question.
We use the same eight-run evaluation protocol as in Q2 and report mean final-answer accuracy.

\paragraph{Findings.}
Across both subsets, Single-Unit Injection yields meaningful gains over the Questions Only variants, while Single-Unit Injection (Question Only) remains low across unit-level and sample-level diagnostics figures.
This gap suggests that the missing information is often the unit answer, rather than merely the decomposition structure.
Moreover, we find that \textbf{injecting a single unit together with its answer can be nearly as effective as providing a full reasoning prefix without answers}, despite conditioning on substantially less context, in both the step-level and the sample-level plots.
This suggests that once a correct intermediate value and statement pair is supplied, the model can proceed with downstream deductions almost as effectively as if it had been guided by many preceding sub-questions.
Equivalently, the limiting factor is not merely knowing what intermediate questions to ask, but reliably deriving \emph{the right intermediate answers} and carrying them forward without drift. Rather than merely indicating that some steps are “critical,” these results suggest that the truly critical signal is often the correct intermediate step answer itself.

\subsection{Reasoning Step Density and Efficiency}

\label{sec:step_length}
\begin{figure*}[t]
    \centering
    \includegraphics[width=\textwidth]{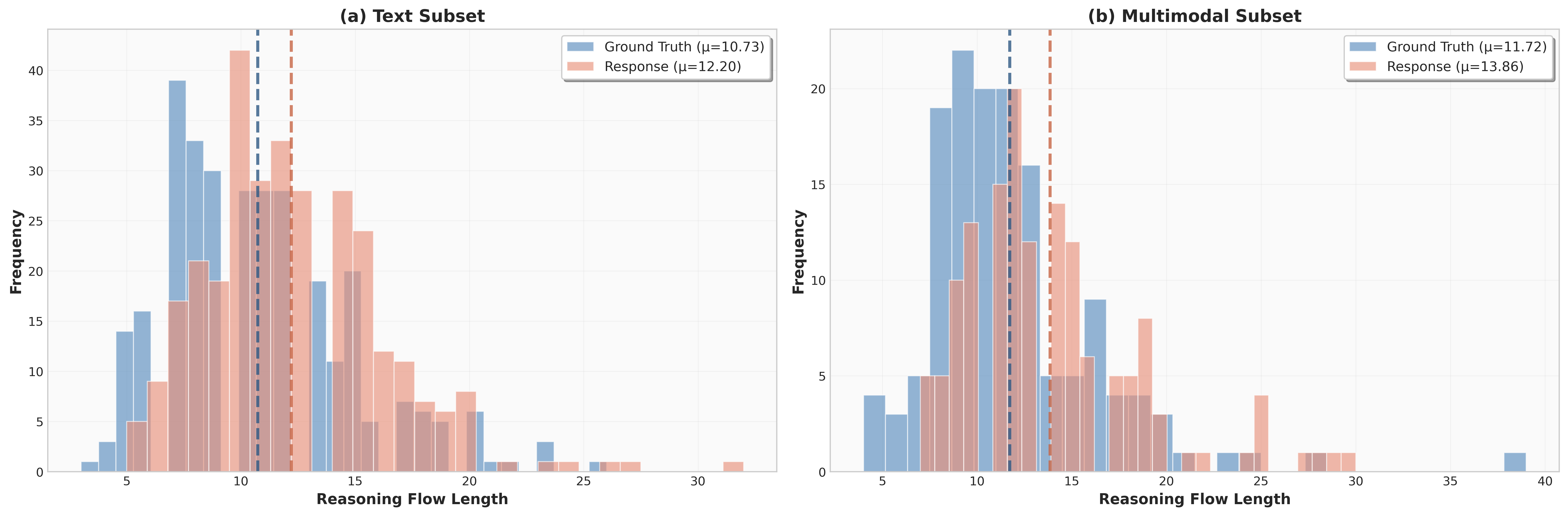}
    \caption{\textbf{Reasoning-flow length distribution.}
    Histograms show the frequency of questions at each length, and dashed vertical lines indicate the corresponding mean lengths.}
    \label{fig:length_distribution_text_mm}
\end{figure*}

Beyond end-to-end accuracy, we also analyze how \emph{efficiently} strong models solve CFE questions across all samples.
Figure~\ref{fig:length_distribution_text_mm} compares the distribution of reasoning-flow lengths between the human-verified ground truth and the model-generated solutions, and Table~\ref{tab:reasoning_length_by_outcome} further stratifies lengths by outcome.

Across both subsets, the reasoning steps of model responses are, on average, longer than those of ground truth, indicating lower step efficiency.
For the text subset, the mean response length is $12.20$ versus a ground-truth mean of $10.73$ (a $+1.47$ step shift; $\approx\!14\%$ longer).
For the multimodal subset, the gap is larger: $13.86$ versus $11.72$ (a $+2.14$ step shift; $\approx\!18\%$ longer).
This rightward shift is visible in Figure~\ref{fig:length_distribution_text_mm} via the separation between the dashed mean markers.
The observed length inflation indicates that current models do not reliably allocate reasoning steps to efficient intermediate quantities.

\begin{table*}[t]
\centering
\scriptsize
\setlength{\tabcolsep}{5pt}
\renewcommand{\arraystretch}{1.15}
\begin{tabular}{p{2.6cm}ccc|ccc}
\toprule
& \multicolumn{3}{c}{\cellcolor{blue!10}\textbf{Text subset} ($n=305$)}
& \multicolumn{3}{c}{\cellcolor{purple!10}\textbf{Multimodal subset} ($n=144$)} \\
\cmidrule(lr){2-4}\cmidrule(lr){5-7}
\textbf{Statistic} & \textbf{Solved} & \textbf{Unsolved} & \textbf{Overall}
& \textbf{Solved} & \textbf{Unsolved} & \textbf{Overall} \\
\midrule
\# Questions
& 179 & 126 & 305
& 74 & 70 & 144 \\
\midrule
GT len.\ (mean$\pm$std)
& $10.60 \pm 4.06$ & $10.91 \pm 4.07$ & $10.73 \pm 4.07$
& $11.04 \pm 3.39$ & $12.44 \pm 5.56$ & $11.72 \pm 4.63$ \\
Resp.\ len.\ (mean$\pm$std)
& $11.91 \pm 3.62$ & $12.62 \pm 4.34$ & $12.20 \pm 3.95$
& $13.19 \pm 4.13$ & $14.57 \pm 5.05$ & $13.86 \pm 4.65$ \\
\bottomrule
\end{tabular}
\caption{\textbf{Reasoning-flow length statistics by outcome.}
We report the lengths of reasoning flows for ground-truth (GT) and model-response (Resp.).}
\label{tab:reasoning_length_by_outcome}
\end{table*}

\subsection{Takeaways and Implications for Stronger Future Models}
\label{sec:takeaways}

Our diagnostic experiments yield three main takeaways about why frontier models underperform on \CFE{}, together with concrete implications for improving future SOTA systems.

\paragraph{(T1) Atomic competence is not the primary bottleneck.}
Under the unit-execution test (Q1), the model attains consistently high accuracy, indicating that many end-to-end failures are \emph{not} driven by missing isolated facts or inability to perform a single local derivation once the correct sub-question is specified.

\paragraph{(T2) Correct intermediate answers are critical.}
Across both text and multimodal subsets, conditions that provide \emph{unit answers} consistently outperform their corresponding ``questions-only'' variants, with the largest gains appearing at mid-range unit indices.
This suggests that the primary bottleneck is not simply identifying a plausible reasoning flow, but reliably \emph{deriving} and \emph{maintaining} correct intermediate states throughout the solution process.
Moreover, injecting a single unit together with its answer can be nearly as effective as providing a much longer reasoning prefix without answers.
Taken together, these results indicate that the truly critical signal is often the \emph{correct intermediate answer itself}: once a key intermediate value or statement is available, it can unlock downstream reasoning and substantially improve end-to-end success.

\paragraph{(T3) Current reasoning is inefficient.}
Models generate longer reasoning flows than the human ground truth in both subsets. This length inflation indicates lower reasoning efficiency, with extra steps creating more opportunities for intermediate drift and error accumulation.

\paragraph{Implications.}
A promising direction is stronger supervision of intermediate states, \eg, step-verified targets, constraint checking, and curricula that reward correct intermediate values rather than only final answers or fluent explanations. 
These findings also motivate hybrid systems that (i) compute or retrieve key intermediate values using stronger tools (\eg, symbolic solvers, verified calculators, or structured retrieval) and (ii) condition the model on these validated intermediates. 
More broadly, improving \CFE{} performance will require more efficient reasoning; training objectives that penalize redundant steps and reward compact derivations are likely to improve both accuracy and efficiency.

\section{Conclusion}
We introduce \CFE{}, a text-and-multimodal benchmark built from commonly used STEM materials, along with a variable-based evaluation protocol that reduces false positives in long-form answer matching. Frontier models still show substantial headroom on \CFE{} across both text-only and multimodal settings. Using a diagnostic framework that constructs reasoning flows from instructor solutions, we find that strong models often execute individual reasoning steps correctly, but fail to reliably derive and preserve correct intermediate states over long derivations. We further observe that the model exhibits lower reasoning efficiency, which creates more opportunities for intermediate errors to accumulate. We hope \CFE{} will serve as a realistic, diagnostic testbed for developing future models, training objectives, and inference strategies that emphasize verifiable intermediate supervision and efficient reasoning.

\subsubsection*{Acknowledgments}
We thank Eric Carlson, Yidong Chong, Jens Jensen, Kevin Zhou, Brian Naranjo, and Ravishankar Sundararaman for their support and valuable feedback throughout this project. We also thank GMI Cloud for providing support for the inference service.

\bibliographystyle{iclr2026_conference}
\bibliography{iclr2026_conference}

\clearpage
\appendix
\section{Variable Value Extraction and Verification Prompts}
\label{apx:prompts}
We present the prompts for extracting and verifying variable values.
\begin{runningexample}[Variable Value Extraction Prompt]
\begin{lstlisting}[
    basicstyle=\footnotesize\ttfamily,
    breaklines=true,
    breakatwhitespace=true,
    columns=flexible
]
You are a strict data extraction engine. Extract values for 
`{variable_list}` from `{response}` based on `{type_list}` and 
'{reference_description}'.

### CRITICAL: SOURCE & CARDINALITY
1. **SOURCE OF TRUTH:** Extract values ONLY from `{response}`.
2. **ONE-TO-ONE MAPPING:** The output list MUST have the exact 
   same number of elements as `{variable_list}`.
3. **ORDER:** Value at index `i` corresponds strictly to 
   variable at index `i`.

### EXTRACTION RULES
- **Numeric:** Extract pure numbers only (remove units).
- **Formula:** Wrap LaTeX in `$` symbols.
- **Other:** Extract exact text with formatting.
- **Missing Data:** If a variable is not found in `{response}`, 
  the value is "null".
    - *Constraint:* Do NOT collapse the list.
    - *Example:* Variables=["x", "y", "z"] where "y" and "z" 
      are missing -> Output=[value_x, "null", "null"]

### INPUT DATA
**Question:** {question}
**Variables:** {variable_list}
**Types:** {type_list}
**Reference description:** {reference_description}
**Response (Extract Values):** {response}

### OUTPUT
Return a valid JSON object:
{
    "short_answer_value_list": [value1, value2, ...]
}
\end{lstlisting}
\end{runningexample}

\begin{runningexample}[Verification Prompt]
\begin{lstlisting}[
    basicstyle=\scriptsize\ttfamily,
    breaklines=true,
    breakatwhitespace=true,
    columns=flexible
]
You are an expert mathematical and logical evaluator. Your task 
is to determine if an extracted answer matches the reference 
(ground truth) answer.

You will be given:
1. `variable_name`: The name of the variable being checked
2. `reference_value`: The ground truth answer (CORRECT)
3. `extracted_value`: The answer extracted from a model's 
   response (TO BE VERIFIED)
4. `question_context` (optional): Context from the original 
   question

### Your Task:

Determine if the `extracted_value` is mathematically/logically 
equivalent to the `reference_value`.

### Evaluation Rules:

**1. Mathematical Equivalence:**
- Check if values are mathematically equal, even if formatted 
  differently
- Examples of CORRECT matches:
  - Reference: 5, Extracted: 5.0 -> CORRECT
  - Reference: "1/2", Extracted: "0.5" -> CORRECT
  - Reference: "50%", Extracted: "0.5" -> CORRECT
  - Reference: "2^2", Extracted: "4" -> CORRECT
  - Reference: "sqrt(16)", Extracted: "4" -> CORRECT

**2. Unit Handling:**
- Values with the same quantity but different units may be 
  correct if equivalent
- Examples:
  - Reference: "1 m", Extracted: "100 cm" -> CORRECT
  - Reference: "1 kg", Extracted: "1000 g" -> CORRECT
  - Reference: "60 mph", Extracted: "96.56 km/h" -> CORRECT 
    (if approximately equal)
- If units are fundamentally different (e.g., meters vs 
  seconds), mark as INCORRECT

...

**7. Special Cases:**
- Mathematical expressions: Check if they simplify to the same 
  value
- Vectors/coordinates: Check component-wise equality
- Sets: Order doesn't matter, content must match
- Fractions: Reduce to simplest form before comparing

### Confidence Levels:

- **high**: Clear match or clear mismatch, no ambiguity
- **medium**: Match with minor unit conversion or rounding 
  involved
- **low**: Uncertain due to formatting ambiguity or complex 
  expressions

### Output Format:

Return a JSON object with:
- `is_correct`: boolean (true if equivalent, false if not)
- `reasoning`: Detailed explanation of your decision
- `confidence`: "high", "medium", or "low"

### Examples:

**Example 1: Exact Match**
Variable: "x"
Reference: 5
Extracted: 5
Output:
{
    "is_correct": true,
    "reasoning": "Both values are exactly 5, perfect match",
    "confidence": "high"
}

...
---

**Variable Name:**
{variable_name}

**Reference Value (Ground Truth):**
{reference_value}

**Extracted Value (To Verify):**
{extracted_value}

**Question Context:**
{question_context}
\end{lstlisting}
\end{runningexample}

\section{Reasoning Flow Construction Prompt}
\label{apx:reason_flow_prompt}
We present the prompt for reasoning flow construction.
\begin{runningexample}[Reasoning Flow Construction Prompt]
\begin{lstlisting}[
    basicstyle=\footnotesize\ttfamily,
    breaklines=true,
    breakatwhitespace=true,
    columns=flexible
]
You are an expert in constructing **traceable and verifiable 
reasoning chains** from reference solutions.

Your task is to decompose the given `Reference Answer` into an 
ordered list of reasoning steps that are **verifiable** and 
**strictly logical**.

The goal is to create a "Chain of Thought" that logically 
reconstructs the answer, where each step acts as a clear 
instruction or question that yields a verifiable result.

---

### Core Principles

1.  **The Q/A (Question/Answer) Principle (CRITICAL):**
    * Think of each step as a Question/Answer pair.
    * `step`: This is the **Question** or **Instruction**. It 
      must pose the problem to be solved (e.g., "What is the 
      formula for X?", "Calculate the value of Y", "Extract 
      the reading from the gauge").
    * `verifiable_answer`: This is the **Solution** or 
      **Result**. It is the concrete answer to the `step`'s 
      instruction (e.g., "F = ma", "9.8", "True", "beaker").

2.  **The Anti-Leaking Rule (CRITICAL):**
    * The **Question** (`step`) **MUST NOT** contain the 
      **Solution** (`verifiable_answer`). The `step` must be 
      formulated in a way that it can be answered *without* 
      already knowing the solution. It must be 
      "answer-agnostic."
    * **VIOLATION (Leaking):** 
      `step: "Identify that the value is 9.8 m/s^2."`
    * **CORRECT (Instruction):** 
      `step: "What is the standard value for gravitational 
      acceleration (g) on Earth?"`
    * **VIOLATION (Leaking):** `step: "Use the value 10N."`
    * **CORRECT (Instruction):** 
      `step: "Extract the value for Force (F) from the 
      Question text."` (Assuming '10N' was in the question)

3.  **The Strict Dependency Rule (Traceability):**
    * Any instruction in a `step` must be solvable using 
      **ONLY** the information from:
        1.  The original `Question` and `Image(s)`.
        2.  The `step` and `verifiable_answer` fields from 
            **previous steps**.
    * You cannot use information from the `Reference Answer` 
      that has not yet been established as a 
      `verifiable_answer` in a prior step.

---

### Input
**Question:**
{question}

**Reference Answer:**
{answer}

---

### Step Requirements
Each object in the `steps` list must contain:

* `step_id`: A 1-based integer for the step order.
* `step`: The instruction or question. This must:
    * Represent a single, explicit **computational**, 
      **logical**, or **information extraction** instruction.
    * Adhere strictly to the **Anti-Leaking Rule**. It must 
      *pose the problem*, not state the answer.
    * Adhere strictly to the **Strict Dependency Rule**.
    * Be an **action or instruction**, not a narrative 
      explanation or a restatement of the answer.
    * Use a dedicated step to extract any new fact, number, 
      or detail from **non-textual inputs (e.g., images, 
      charts)**.
* `verifiable_answer`: The result. This must:
    * Be the single, deterministic, and objective 
      **solution** to the `step` instruction.
    * Be a **formula** (e.g., `a = F / m`), a **number** 
      (e.g., `5.0`), a **boolean** (e.g., `True`), or a 
      **short string literal** (e.g., `beaker`).

---

### Output Format (JSON Only)
Return a JSON object with:
- `sub_steps`: ordered list of objects, each containing:
  - `sub_step_id`: integer (1-based order)
  - `sub_step`: the instruction or question for an 
    intermediate reasoning sub-step
  - `verifiable_answer`: the **formula**, **number**, or 
    **boolean** answer to the instruction or question

---

### Example

{
  "steps": [
    {
      "step_id": 1,
      "step": "What is the type of glassware shown in the 
               image?",
      "verifiable_answer": "beaker"
    },
    {
      "step_id": 2,
      "step": "State the formula for Newton's Second Law.",
      "verifiable_answer": "F = m * a"
    },
    {
      "step_id": 3,
      "step": "Rearrange the formula from Step 2 
               (F = m * a) to solve for acceleration (a).",
      "verifiable_answer": "a = F / m"
    },
    {
      "step_id": 4,
      "step": "Substitute F=10N (from Question) and m=2kg 
               (from Question) into the formula from Step 3 
               (a = F / m), and calculate the result.",
      "verifiable_answer": 5.0
    }
  ]
}

Return **only the JSON object** described above and keep 
latex notation.
\end{lstlisting}
\end{runningexample}

\section{Reasoning Flow Example}
We show a reasoning flow example in Table~\ref{tab:reasoning_flow}.

\begin{longtable}{c L{5.8cm} L{7.5cm}}
\caption{An example reasoning flow for a representative problem. Given a question, the reference solution is decomposed into an ordered sequence of atomic, verifiable reasoning units $R = [u_1, u_2, \dots, u_n]$. Each unit $u_i = \langle u_i^q, u_i^a \rangle$ pairs a sub-question $u_i^q$ with its objectively checkable target answer $u_i^a$.}
\label{tab:reasoning_flow} \\
\toprule
\multicolumn{3}{L{15cm}}{\textbf{Question:} Short-circuited parallel plate electrodes of area $A$ enclose a lossy dielectric of thickness $s$ with dielectric permittivity $\varepsilon$ and ohmic conductivity $\sigma$. The lossy dielectric at time $t=0$ has a uniformly distributed free volume charge density $\rho_{0}$. Neglect fringing field effects. What is the current $i(t)$ flowing through the short circuit?} \\
\midrule
\textbf{Step} & \textbf{Question $u^q_i$} & \textbf{Answer $u^a_i$} \\
\midrule
\endfirsthead
\caption[]{(continued)} \\
\toprule
\textbf{Step} & \textbf{Question $u^q_i$} & \textbf{Answer $u^a_i$} \\
\midrule
\endhead
\bottomrule
\endlastfoot

$u^q_1$ / $u^a_1$ &
Write the expression for the total current density in a lossy dielectric in terms of $\sigma$, $\varepsilon$, and $E_x$ at the electrode ($x = s$). &
$\displaystyle \frac{i(t)}{A} = \sigma E_x\big|_{x=s} + \varepsilon \frac{\partial E_x}{\partial t}\bigg|_{x=s}$ \\
\midrule

$u^q_2$ / $u^a_2$ &
For a uniform free volume charge density $\rho_f(t)$ inside the slab of thickness $s$, state the relation between $E_x(x{=}s)$ and $\rho_f(t)$ using Gauss's law. &
$\displaystyle E_x(x{=}s) = \frac{\rho_f(t)\, s}{2\varepsilon}$ \\
\midrule

$u^q_3$ / $u^a_3$ &
Differentiate the expression from Step~2 with respect to time to obtain $\partial E_x/\partial t$ at $x = s$. &
$\displaystyle \left.\frac{\partial E_x}{\partial t}\right|_{x=s} = \frac{s}{2\varepsilon}\,\frac{\partial \rho_f}{\partial t}$ \\
\midrule

$u^q_4$ / $u^a_4$ &
Substitute the expressions for $E_x(x{=}s)$ and $\partial E_x/\partial t$ from Steps~2 and~3 into the current density expression from Step~1. &
$\displaystyle \frac{i(t)}{A} = \sigma\,\frac{\rho_f\, s}{2\varepsilon} + \varepsilon\,\frac{s}{2\varepsilon}\,\frac{\partial \rho_f}{\partial t}$ \\
\midrule

$u^q_5$ / $u^a_5$ &
Simplify the expression from Step~4 algebraically. &
$\displaystyle \frac{i(t)}{A} = \frac{\sigma\, s}{2\varepsilon}\,\rho_f + \frac{s}{2}\,\frac{\partial \rho_f}{\partial t}$ \\
\midrule

$u^q_6$ / $u^a_6$ &
State the time-dependence of $\rho_f(t)$ for initial value $\rho_0$ in a lossy dielectric with relaxation time $\tau$. &
$\displaystyle \rho_f(t) = \rho_0\, e^{-t/\tau}$ \\
\midrule

$u^q_7$ / $u^a_7$ &
Express the relaxation time $\tau$ in terms of $\varepsilon$ and $\sigma$. &
$\displaystyle \tau = \frac{\varepsilon}{\sigma}$ \\
\midrule

$u^q_8$ / $u^a_8$ &
Compute $\partial \rho_f/\partial t$ from Step~6 using $\tau$ from Step~7. &
$\displaystyle \frac{\partial \rho_f}{\partial t} = -\frac{\sigma}{\varepsilon}\,\rho_0\, e^{-t/\tau}$ \\
\midrule

$u^q_9$ / $u^a_9$ &
Substitute $\rho_f(t)$ and $\partial \rho_f/\partial t$ from Steps~6 and~8 into the expression from Step~5. &
$\displaystyle \frac{i(t)}{A} = \frac{\sigma\, s}{2\varepsilon}\,\rho_0\, e^{-t/\tau} + \frac{s}{2}\!\left(-\frac{\sigma}{\varepsilon}\,\rho_0\, e^{-t/\tau}\right)$ \\
\midrule

$u^q_{10}$ / $u^a_{10}$ &
Algebraically combine the two terms from Step~9. &
$\displaystyle \frac{i(t)}{A} = 0$ \\
\midrule

$u^q_{11}$ / $u^a_{11}$ &
State the resulting total current $i(t)$ through the short circuit. &
$\displaystyle i(t) = 0$ \\

\end{longtable}

\end{document}